\documentclass{Interspeech2024}
\usepackage{hyperref}

\usepackage{xcolor}

\interspeechcameraready

\title{LearnerVoice: A Dataset of Non-Native English Learners’ Spontaneous Speech}

\name[affiliation={1,2}]{Haechan}{Kim}
\name[affiliation={2}]{Junho}{Myung}
\name[affiliation={2}]{Seoyoung}{Kim}
\name[affiliation={1}]{Sungpah}{Lee}
\name[affiliation={3}]{Dongyeop}{Kang}
\name[affiliation={1,2}]{Juho}{Kim}

\address{
  $^1$Ringle, Republic of Korea\\
  $^2$School of Computing, KAIST, Republic of Korea \\
  $^3$University of Minnesota, US}
\email{\{kim.haechan, junho00211, youthskim, juhokim\}@kaist.ac.kr, sungpah@ringleplus.com, dongyeop@umn.edu}


\begin{document}

\maketitle
    
\begin{abstract}
Prevalent ungrammatical expressions and disfluencies in spontaneous speech from second language (L2) learners pose unique challenges to Automatic Speech Recognition (ASR) systems. However, few datasets are tailored to L2 learner speech. We publicly release LearnerVoice, a dataset consisting of 50.04 hours of audio and transcriptions of L2 learners' spontaneous speech. Our linguistic analysis reveals that transcriptions in our dataset contain L2S (L2 learner's Spontaneous speech) features, consisting of ungrammatical expressions and disfluencies (e.g., filler words, word repetitions, self-repairs, false starts), significantly more than native speech datasets. Fine-tuning whisper-small.en with LearnerVoice achieves a WER of 10.26\%, 44.2\% lower than vanilla whisper-small.en. Furthermore, our qualitative analysis indicates that 54.2\% of errors from the vanilla model on LearnerVoice are attributable to L2S features, with 48.1\% of them being reduced in the fine-tuned model.
\end{abstract}
\section{Introduction}

Spontaneous speech is often characterized by disfluencies such as filler words, self-repairs, word repetitions, and false starts, which are not commonly encountered in read speech \cite{dufour2009spontaneous, boulademareuil05_diss}. Second language (L2) learners exhibit these disfluencies more frequently in spontaneous speech, along with ungrammatical utterances not typically found in native speakers' speech. These ungrammatical expressions and disfluencies, which we define as \textbf{L2S} (\textbf{L2}'s \textbf{S}pontaneous speech) features, increase the complexity of the Automatic Speech Recognition (ASR) task \cite{qiao21b_interspeech}.

The precise transcription of L2S features stands out as a crucial element in the automatic assessment of speaking tests for L2 learners \cite{hassanali2015automatic, gretter2019automatic}. One of the commonly used as an evaluation scheme of L2 speaker's speaking abilities is Complexity, Accuracy, and Fluency (CAF) triad \cite{housen2012complexity}. Ungrammatical utterances contribute to accuracy, while disfluencies are pivotal components for evaluating the fluency \cite{yan2018complexity}. Despite the importance of the accurate transcription of L2S features, recent ASR systems show a higher error rate in transcribing the disfluencies \cite{qiao21b_interspeech}, and often automatically rectify the grammatical errors \cite{knill18_interspeech}. One of the main reasons for such challenges comes from the lack of publicly available datasets or benchmarks that comprehensively encompass the L2S features in a spontaneous L2 speech \cite{qiao21b_interspeech}.
 
To this end, we construct and publicly release LearnerVoice\footnote{Our dataset is available at: https://prep.ringleplus.com/research}, a spontaneous English speech dataset collected from L2 learners. LearnerVoice consists of a fully transcribed 50.04 hours of audio (229,671 tokens) spoken by 58 L2 English learners who have Korean as their first language. The audio was collected from Ringle\footnote{https://www.ringleplus.com/}, an online learning platform that hosts one-to-one video tutoring sessions between L2 learners and native English tutors. To accurately capture L2S features in the dataset, we recruited annotators who can fully understand the biased accent from L2 learners. Then we trained them on what L2S features are and how they should be transcribed by providing case examples. Our analysis result shows that our dataset includes significantly more L2S features than Switchboard \cite{godfrey1992switchboard} and Librispeech \cite{panayotov2015librispeech}, popular native speech datasets.

To further investigate the importance of considering L2S features in improving ASR performance, we also identify a taxonomy with 9 types of errors made by the ASR model for L2 learners based on previous works \cite{goldwater2010words, zhu2023transcription, ZAFAR2004719, spilker2000processing, kim2009error, strik11_slate, frieske2024hallucinations, rodrigues2019analyzing, vasilescu2011cross}. Then we asked annotators to tag the error types on the transcription inferred from the subset of LearnerVoice by vanilla whisper-small.en, a state-of-the-art ASR model \cite{radford2023robust}. Results suggest that 54.2\% of the errors are tagged to the L2S features related types (Filler word, Self-repair, and Ungrammatical Expression). 

Fine-tuning whisper-small.en model with LearnerVoice show a WER of 10.26\%, which corresponds to a 44.2\% decrease in WER compared to that of vanilla whisper-small.en. Additionally, we measured the change rates of each error type from both the vanilla model and the fine-tuning model by error tagging. As a result, there was a reduction of 48.1\% for error types associated with L2S features, while errors related to non-L2S features only decreased by 19.2\%. We believe our work will serve as a cornerstone of future research in ASR to be more inclusive by supporting diverse L2 learners with different L2S feature distributions. 

The contributions of this paper are as follows:
\begin{itemize}[leftmargin=2em]
    \item Public release of LearnerVoice, a spontaneous English speech dataset containing 50.04 hours of audio and corresponding transcription collected from L2 learners.

    \item Identification of L2S features, consisting of ungrammatical expressions and disfluencies that frequently appear in L2 learners' spontaneous speech. Experimental results show the influence of the L2S features on ASR errors for L2 learners' spontaneous speech.
\end{itemize}
\section{LearnerVoice Dataset}

\subsection{Dataset Overview}
Audio is collected through Ringle, a platform based in Korea, which matches non-native English learners with native tutors for one-to-one tutoring. Lessons are conducted via video calls for either 20 or 40 minutes with various topics such as daily life, business/economics, current affairs/politics, and culture/sports. The audio from 239 lessons contains spontaneous speech from L2 learners. Users were informed and provided consent that lesson data could be released as a public dataset.

As a result, LearnerVoice consists of a total of 50.04 hours of audio, which is obtained from 239 lessons, with a sampling rate of 32,000Hz. The lessons consist of 168 (70.3\%) 20-minute lessons and 71 (29.7\%) 40-minute lessons. The transcription of LearnerVoice contains a total of 229,671 tokens based on whitespace tokenization. Compared to the existing native speaker spontaneous speech datasets, the number of tokens per hour is lower (the subset of Switchboard has 228,107 tokens in 22.92 hours). This is attributed to the slower speaking rate of L2 learners. 

\subsection{Dataset Construction}
The voices of learners and tutors are collected separately as individual channels. To prepare the audio in a trainable format, we segmented the audio into short units using Voice Activity Detection model released by pyannote \cite{Bredin2021}. Subsequently, the segmented audio was provided to human annotators for transcribing.

\begin{figure}[htb!]
    \centering
    \includegraphics[width=0.80\linewidth]{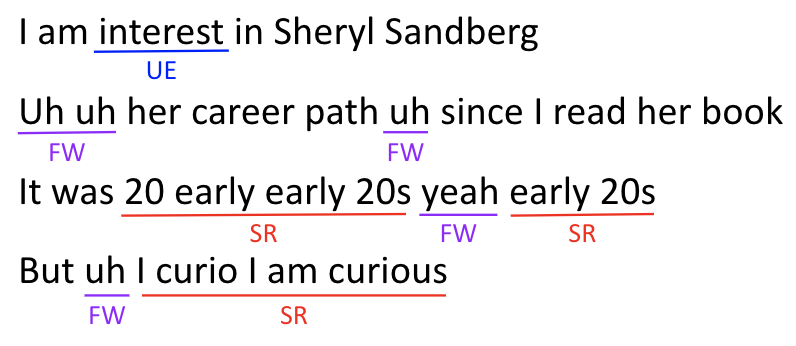}
    \vspace{-.3cm}
    \caption{Examples of L2S features from the transcriptions of LearnerVoice, where filler words (FW), self-repairs (SR), and ungrammatical expressions (UE) are well represented. Word repetitions, self-repair, fragment, and false starts are all considered as types of self-repair and are labeled as SR.}
    \label{fig:transcription_example}
\end{figure}

We recruited annotators who have resided in the United States for over a year or have TOEFL scores exceeding 100 through an online recruitment posting. Recognizing the learner's accent is an important qualification for annotators, so we selected native Korean for the task. The most important consideration during the construction of our dataset was ensuring that the L2S features present in the audio were accurately reflected in the transcription. Thus, we trained annotators on what L2S features are and how they should be transcribed by providing audio-transcription paired examples. Figure~\ref{fig:transcription_example} is an example of our transcription where L2S features are accurately reflected.
\vspace{-.1cm}
\vspace{-.5cm}

\subsection{L2 Learners Distribution}
The dataset comprises speech recordings from 58 L2 learners whose first language is Korean. It consists of 38 female and 20 male learners with ages ranging from 20s to 40s. Since many did not have official English-speaking test scores, we utilized the CAF engine developed by Ringle to report the distribution of English-speaking proficiency of learners. The CAF engine measures Complexity, Accuracy, and Fluency of speech based on previous research \cite{yan2018complexity, housen2012complexity}. Subsequently, it assigns a level to speakers based on their IELTS\footnote{https://ielts.org/} speaking band (ranging from 1 to 9). The CAF engine showed root mean square error between the predicted bands and the ground truth bands as 0.66 out of 9 when tested on 98 lessons with ground truth bands. Table~\ref{tab:CAF_score} presents the prediction of the CAF engine for the 58 L2 learners. The band predictions were derived from the most recent 30 lessons of each learner, some of which may not be included in the dataset. They are distributed across IELTS bands 4 to 9, with average score of 5.95. This value is similar to the average IELTS speaking band score of 5.9 for Koreans in 2022~\cite{IELTS}.

\begin{table}[htb!]
    \centering
    \small
    \begin{tabular}{l|cccccc} 
        \hline
        Band & 4 & 5 & 6 & 7 & 8 & 9\\ 
        \hline
        Complexity & 14\% & 36\% & 21\% & 17\% & 9\% & 3\%\\
        Accuracy & 21\% & 16\% & 24\% & 24\% & 9\% & 7\%\\
        Fluency & 2\% & 31\% & 31\% & 28\% & 5\% & 3\%\\
        \hline
    \end{tabular}
    \caption{The percentage of number of L2 learners categorized by IELTS band inferred through the CAF engine. The percentage has been rounded to the nearest whole number. Learners are mostly distributed across bands 4, 5, 6, and 7.}
    \label{tab:CAF_score}
\end{table}
\vspace{-0.6cm}



\subsection{L2S Features Distribution}

We define L2S features as distinct speech characteristics of L2 learners that prominently appear in a spontaneous speech. We assess and compare the occurrences of L2S features in LearnerVoice with those in other representative speech datasets from native speakers. We selected Switchboard \cite{godfrey1992switchboard} and Librispeech \cite{panayotov2015librispeech} as they exemplify native-spontaneous and native-read speech, respectively. To ensure a fair comparison, we then randomly selected subsets of Switchboard and Librispeech, each containing 228,107 tokens and 229,026 tokens, respectively.


\subsubsection{Quantifying L2S Features in the Dataset}
The quantification of L2S features is represented by filler words per token, self-repairs per token, and grammatical errors per c-unit. The concept of token is based on whitespace tokenization. A c-unit (communication unit) is defined as "an independent clause with its modifiers," often serving as the basic unit for measuring grammatical errors \cite{Eisenberg2018PercentGR}. The detailed quantifying process is as follows: (1) Firstly, filler words were counted using hard-coded detection. (2) Self-repairs were identified after removing filler words from the original sentences. Self-repairs were defined as repetitions of lemma n-grams, where n is less than 5. (4) Next, to find grammatical errors, sentences with filler words and self-repairs removed were inputted into CoEdIT-large~\cite{raheja2023coedit}, along with the prompt "Fix grammatical errors in this sentence." Then, ERRANT~\cite{bryant-etal-2017-automatic} was employed to count the number of grammatical errors in the input sentences. 

\subsubsection{Comparison with Other Datasets}

\begin{figure}[htb!]
    \centering
    \includegraphics[width=1\linewidth]{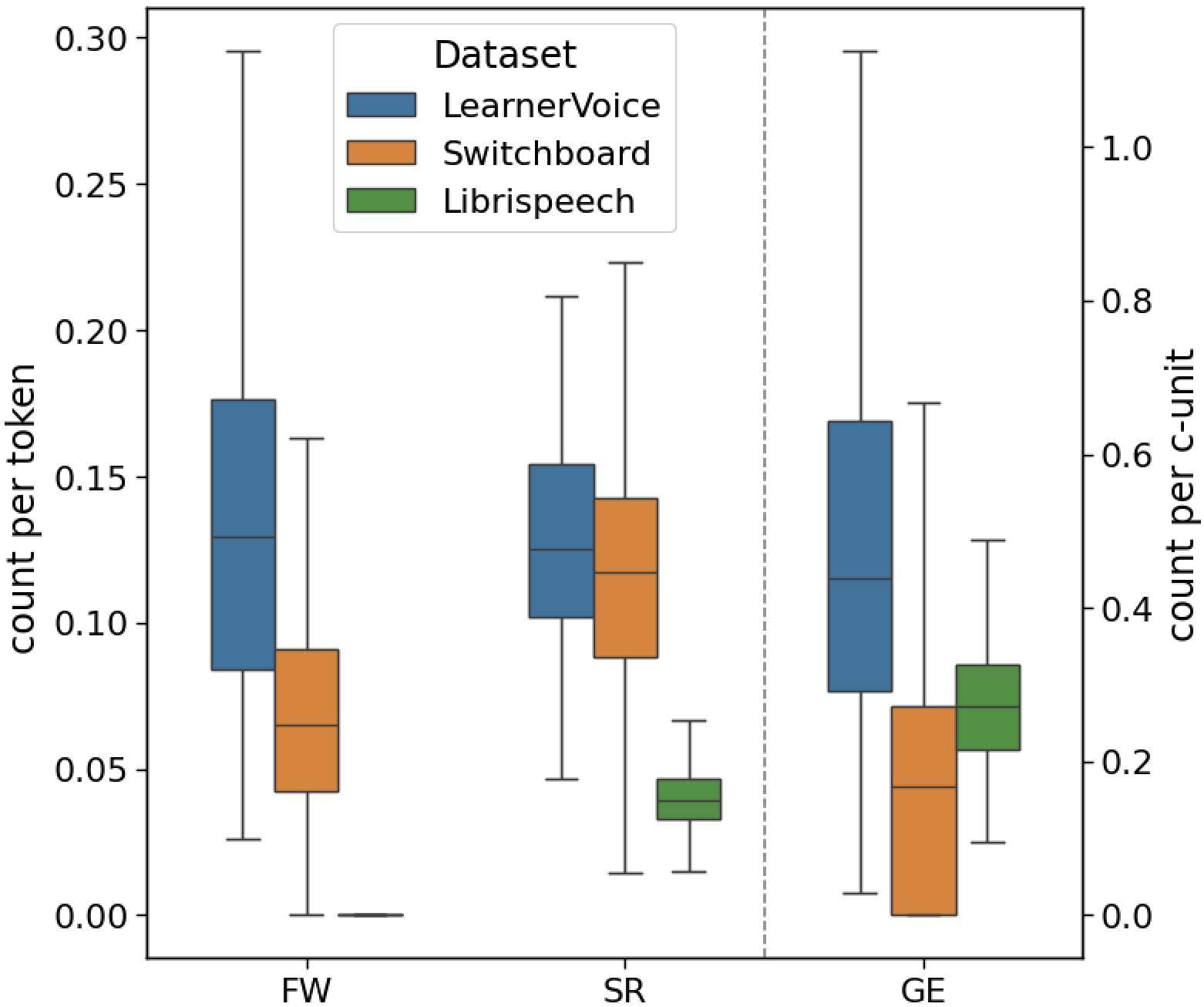}
    \caption{Distribution of filler words per token (FW), self-repairs per token (SC) and grammatical errors per c-unit (GE) by datasets. FW and SC are followed the left-hand side y-axis and GE is followed the right-hand side y-axis. This statistically significant difference in the abundance of L2S features in LearnerVoice compared to Switchboard and Librispeech.}
    \label{fig:L2S_features}
\end{figure}


\noindent 
Figure~\ref{fig:L2S_features} shows the distributions of L2S features for each dataset. The Mann-Whitney U test demonstrates that for all L2S features, LearnerVoice exhibits significant differences compared to both Switchboard and Librispeech.

The frequency of filler words is significantly higher in LearnerVoice (N=239, $.136 \pm .063$) than in Switchboard (N=573, $.077 \pm .077$) ($p < .001$) and in Librispeech (N=240, $.000 \pm .000$) ($p < .001$). In case of self-repairs, it is also significantly higher in LearnerVoice ($.127 \pm .037$) than in Switchboard ($.118 \pm .049$) ($p < .01$) and in Librispeech ($.040 \pm .012$) ($p < .001$). Furthermore, the ratio of grammar errors was higher in LearnerVoice ($.477 \pm .248$) compared to Switchboard ($.200 \pm .210$) ($p < .001$) and Librispeech ($.280 \pm .091$) ($p < .001$). These results suggest that our dataset contains more L2S features than other datasets.
\section{Fine-tuning with LearnerVoice}
To evaluate whether LearnerVoice can enhance ASR performance for L2 spontaneous speech compared to existing datasets, we fine-tuned an ASR model using both LearnerVoice and a comparable subset of a native spontaneous speech dataset, Switchboard. 

\vspace{-0.05cm}
\subsection{Dataset Used for Fine-Tuning}
We chose the dataset to be used for fine-tuning based on the number of tokens in the transcriptions for comparison. LearnerVoice was divided into a training set of 185,463 tokens (41.12 hours) and a validation set of 20,705 tokens (4.50 hours). For Switchboard, a training set of 185,475 tokens (18.46 hours) and a validation set of 18,562 tokens (2.00 hours) were used. Each model fine-tuned with the respective datasets was evaluated on the LearnerVoice test set of 23,503 tokens (4.42 hours) and Switchboard test set of 24,070 tokens (2.45 hours).

\vspace{-0.05cm}
\subsection{Experiment Setting}
The baseline ASR model used was whisper-small.en model released by OpenAI \cite{radford2023robust}. The whisper-small.en model has 244 million parameters and is capable of inference only in English. We selected the whisper-small.en model as the baseline model because multilingual inference models often misrecognized Korean English pronunciation as other languages (e.g., Korean, Japanese). 
The AdamW criterion was employed with a starting learning rate of 1e-5, with the first 500 training steps used as warm-up steps. To expedite training, gradient accumulation was applied every 2 steps.

\vspace{-0.05cm}
\subsection{Experiment Result}

\noindent 
Table~\ref{tab:fine-tuning-result} displays the WERs for the LearnerVoice test set resulting from the vanilla whisper-large-v3 and vanilla whisper-small.en models, as well as the whisper-small.en model fine-tuned with LearnerVoice and Switchboard. The model chosen for testing was based on the validation loss measured during training. For LearnerVoice, a model trained for 2.04 epochs was selected, while for Switchboard, a model trained for 1.81 epochs was chosen.

\begin{table}[htb!]
    \centering
    \small
    \begin{tabular}{lll|c} 
        \hline
        Model & train & test &  WER\\ 
        \hline
        large-v3 & - & LearnerVoice  & 19.18\\
        small.en & - & LearnerVoice  & 18.38\\
        small.en & Switchboard & LearnerVoice  & 15.03\\
        small.en & LearnerVoice & LearnerVoice  & \textbf{10.26}\\
        \hline
        small.en & - & Switchboard  & 20.18\\
        small.en & LearnerVoice & Switchboard  & \textbf{19.01}\\
        \hline
    \end{tabular}
    \caption{WER of fine-tuning with LearnerVoice and Switchboard}
    \label{tab:fine-tuning-result}
\end{table}
\vspace{-0.45cm}

The WERs observed for the vanilla whisper-large-v3 model and whisper-small.en model were 19.18\% and 18.38\%, respectively. The inferior performance of the whisper-large-v3 model can be attributed to its multilingual nature, which may result in biased accents from L2 speakers being recognized as different languages. After fine-tuning the whisper-small.en model with the same amount of data from LearnerVoice and Switchboard, the WERs were observed to be 10.26\% and 15.03\%, respectively, on the LearnerVoice test set. This indicates that the model fine-tuned with LearnerVoice experienced a 44.2\% decrease in WER compared to the best-performing model before fine-tuning. Furthermore, when Switchboard was used as the test set, the WERs for the vanilla whisper-small.en and the fine-tuned model with LearnerVoice were 20.18\% and 19.01\%, respectively. This suggests that the model trained with LearnerVoice shows a slight improvement in ASR performance even for native spontaneous speech.
\section{ASR Error Tagging on L2 Speech}

To understand the importance of L2S features in ASR for L2 learners, we analyzed the causes of ASR errors on L2 learners' speech using (1) vanilla whisper-small.en model and (2) whisper-small.en model after fine-tuning with LearnerVoice.

\subsection{ASR Error Taxonomy}
\begin{table*}[t!]
    \centering
    \small
    \resizebox{\textwidth}{!}{
    \begin{tabular}{l|l|l} 
        \toprule
        \multicolumn{1}{c|}{\textbf{Category}} & \multicolumn{1}{c|}{\textbf{Description}} & \multicolumn{1}{c}{\textbf{Example}} \\ 
        \midrule \midrule
        Filler Word (FW) & \begin{tabular}[t]{@{}l@{}} Errors caused by filler words spoken by the speaker \cite{goldwater2010words, zhu2023transcription}\\ (e.g. uh, huh, yeah, oh, hmm, um, etc.) \end{tabular} & \begin{tabular}[t]{@{}l@{}} REF: yeah	ah	yeah	um	very\\OUT: yeah	**	yeah	**	very \end{tabular} \\[0.5cm]
        
        Self-repair (SR) & \begin{tabular}[t]{@{}l@{}} Errors attributable to word repetition, self-repair, or fragment \cite{ZAFAR2004719, spilker2000processing} \end{tabular} & \begin{tabular}[t]{@{}l@{}} REF: it	can	it can increase\\OUT: it can ** *** increase \end{tabular} \\[0.5cm]
        
        Ungrammatical Expression (UE) & \begin{tabular}[t]{@{}l@{}} Errors caused by ASR models auto-correcting the grammatical error\\ spoken by the speaker \cite{kim2009error, strik11_slate} \end{tabular} & \begin{tabular}[t]{@{}l@{}} REF: i seems to be affected\\OUT: i seem to be affected\end{tabular}\\[0.5cm]
        
        Pronunciation/Accent (PA) & \begin{tabular}[t]{@{}l@{}} Errors caused by slips in how the speaker pronounced the words \cite{ZAFAR2004719} \end{tabular} & \begin{tabular}[t]{@{}l@{}} REF: i am worker\\OUT: i am walker\end{tabular} \\[0.5cm]
        
        Hallucination (HL) & \begin{tabular}[t]{@{}l@{}} Errors that are nonsensical or unfaithful to the provided source content \cite{frieske2024hallucinations}\end{tabular} & \begin{tabular}[t]{@{}l@{}} REF: *** free free talking\\OUT: are you already talking \end{tabular} \\[0.5cm]
        
        Dictionary Error (DE) & \begin{tabular}[t]{@{}l@{}} Errors caused by words that are missing from the dictionary \cite{ZAFAR2004719}\\(e.g. proper nouns) \end{tabular} & \begin{tabular}[t]{@{}l@{}} REF: article built with sheryl sandberg\\OUT: article built with sheraton burr\end{tabular} \\[0.5cm]
        
        Ambient Noise/Audio Quality (AA) & \begin{tabular}[t]{@{}l@{}} Errors caused by low audio quality \cite{ZAFAR2004719, rodrigues2019analyzing} (e.g. ambient noise) \end{tabular} & -- \\[0.3cm]
        
        Homophone (HP) & \begin{tabular}[t]{@{}l@{}} Errors caused by homophones \cite{vasilescu2011cross} \end{tabular} & \begin{tabular}[t]{@{}l@{}} REF: is too big\\OUT: is two big\end{tabular} \\[0.5cm]
        
        Miscellaneous (MC) & \begin{tabular}[t]{@{}l@{}} Errors that do not fall into the categories above (e.g. two interchangeable\\spellings of the same word, misalignment in REF and OUT due to inaccuracy\\ in the Levenshtein \cite{levenshtein1966binary} algorithm, etc.) \end{tabular} & \begin{tabular}[t]{@{}l@{}} REF: okay okay \\OUT: ok ok \end{tabular} \\
        \bottomrule
    \end{tabular}}
    \caption{Definitions and examples of nine ASR error types. REF and OUT are the ground truth text from the transcription of LearnerVoice and the output text from the model, respectively. `*' is used for visually aligning the two texts.}
    \label{tab:error_taxonomy_definition}
    \vspace{-0.5cm}
\end{table*}

We referred to literature \cite{goldwater2010words, zhu2023transcription, ZAFAR2004719, spilker2000processing, kim2009error, strik11_slate, frieske2024hallucinations, rodrigues2019analyzing, vasilescu2011cross} on addressing the causes of ASR difficulties to identify nine error types. Table~\ref{tab:error_taxonomy_definition} presents the definitions and examples of them.

\subsection{Methods}
To see whether LearnerVoice effectively influences error types related to L2S features, we compared the error type distribution before and after fine-tuning. For this, we sampled 16 out of the total of 239 lessons that comprise LearnerVoice, which represents 6.7\% of the entire dataset. Then we tagged errors based on the ASR Error Taxonomy in Table~\ref{tab:error_taxonomy_definition} that could be found in the text derived from the (1) vanilla whisper-small.en model and (2) whisper-small.en model fine-tuned with LearnerVoice to analyze the causes of ASR errors.

For annotation, we recruited annotators with high proficiency in English and familiarity with the typical Korean English accents. Annotators underwent a 30-minute on-boarding session explaining the task and error definitions, followed by another 30 minutes of working with the authors on example lessons. To ensure the robustness of the annotation, two annotators independently worked on each lesson and inter-rater reliability was calculated.

\subsection{Results}
\vspace{-0.4cm}
\begin{figure}[htb!]
    \centering
    \includegraphics[width=1\linewidth]{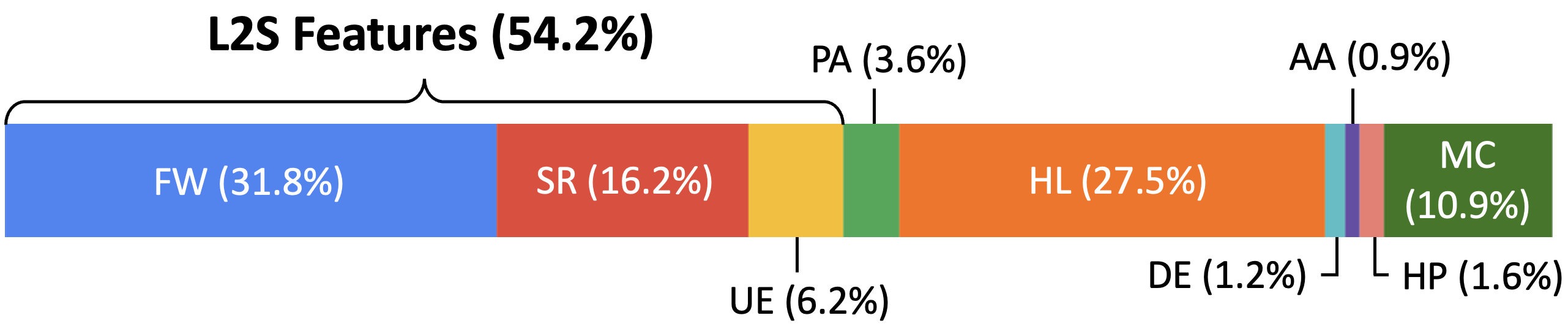}
    \vspace{-0.5cm}
    \caption{The distribution of error types for the vanilla whisper-small.en model indicates that ASR errors are predominantly influenced by error types related to L2S features. The error tags are represented in abbreviated form as shown in Table~\ref{tab:error_taxonomy_definition}.}
    \label{fig:error_distribution_for_vanilla_whisper_large_v3}
\end{figure}

\noindent
Inter-rater reliability (Cohen's kappa score) for the collected tags was 0.77. In the results for the vanilla whisper-small.en (Figure~\ref{fig:error_distribution_for_vanilla_whisper_large_v3}), a total of 4139 errors were observed, and 54.2\% of them are associated with the L2S features. Filler word, Self-repair, and Ungrammatical Expression account for 37.6\%, 17.1\%, and 4.8\% of all errors, respectively. 

\begin{figure}[htb!]
    \centering
    \includegraphics[width=1\linewidth]{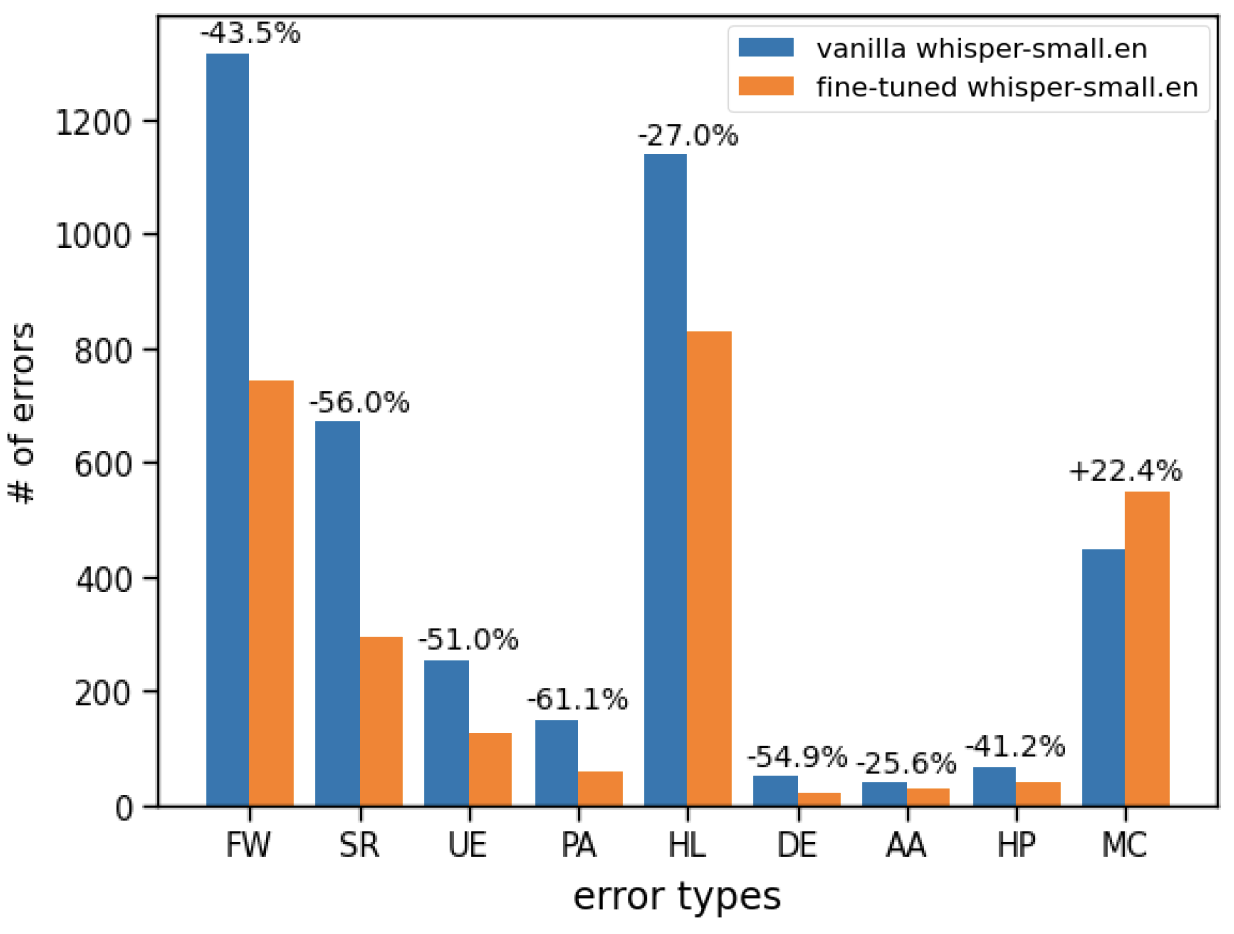}
    \vspace{-0.5cm}
    \caption{Change ratio of error counts by error types for the vanilla whisper-small.en and whisper-small.en fine-tuned by LearnerVoice. There is a much greater reduction in errors for error types related to L2S features compared to error types that are not related.}
    \label{fig:error_tag_comparison}
\end{figure}

Figure~\ref{fig:error_tag_comparison} presents a comparison between the results of the vanilla model and the fine-tuned model. Upon examining the change ratio across different error types, we observe the highest error reduction rates for Pronunciation/Accent, Self-repair, Dictionary Error, Ungrammatical Utterance, and Filler word, in order. The decrease in Dictionary Error seems to be associated with errors in proper nouns linked to platform names. Excluding this, the results indicate a 48.1\% decrease in errors stemming from L2S features after model fine-tuning. Conversely, errors originating from non-L2S features decreased by 19.2\%. This means that LearnerVoice effectively captures L2S features, which aids in addressing L2 spontaneous speech in ASR. Furthermore, the notably high error reduction rate of Pronunciation/Accent suggests that the fine-tuned model effectively accommodates non-native accents.
\section{Conclusion}
We present LearnerVoice, a dataset consisting of 50.04 hours of audio and transcriptions of L2 learners' spontaneous speech. Our linguistic analysis reveals that transcriptions in our dataset contain L2S (L2 learner's Spontaneous speech) features---consisting of ungrammatical expressions and disfluencies (filler words, word repetitions, self-repairs, and false starts)---significantly more than other native speech datasets. Through fine-tuning the whisper-small.en model with LearnerVoice, we achieve a notable reduction in Word Error Rate (WER) compared to that of the vanilla model.
Additionally, our ASR error tagging analysis uncovers that a considerable portion of ASR errors stem from the L2S features, underscoring the importance of addressing the L2S features in ASR systems. We anticipate that LearnerVoice and the insights regarding the impact of the L2S features in ASR systems for L2 learners' spontaneous speech will serve as a cornerstone for future research.









\bibliographystyle{IEEEtran}
\bibliography{mybib}

\end{document}